\documentclass[10pt, a4paper]{article}
\usepackage[final]{lrec2026} 
\usepackage{booktabs}
\usepackage{multirow} 
\usepackage{amsmath} 
\usepackage{tcolorbox}\usepackage{tabularx}
\usepackage{adjustbox}
\usepackage{graphicx}
\usepackage{array}
\title{Towards Consistent Detection of Cognitive Distortions: LLM-Based Annotation and Dataset-Agnostic Evaluation\thanks{Accepted at LREC-2026}}

\name{Neha Sharma, Navneet Agarwal, Kairit Sirts} 

\address{University of Tartu, Estonia \\
         \texttt{\{neha.sharma, navneet.agarwal, kairit.sirts\}@ut.ee}
         }

\abstract{
Text-based automated Cognitive Distortion detection is a challenging task due to its subjective nature, with low agreement scores observed even among expert human annotators, leading to unreliable annotations.  
We explore the use of Large Language Models (LLMs) as consistent and reliable annotators, and propose that multiple independent LLM runs can reveal stable labeling patterns despite the inherent subjectivity of the task.
Furthermore, to fairly compare models trained on datasets with different characteristics, we introduce a dataset-agnostic evaluation framework using Cohen's kappa as an effect size measure. This methodology allows for fair cross-dataset and cross-study comparisons where traditional metrics like F1 score fall short. Our results show that GPT-4 can produce consistent annotations (Fleiss's Kappa = 0.78), resulting in improved test set performance for models trained on these annotations compared to those trained on human-labeled data.
While human expert verification was inconclusive on our target dataset, our findings suggest that LLMs can offer a scalable and internally consistent alternative for generating training data that supports strong downstream performance in subjective NLP tasks.
 \\ \newline \Keywords{Cognitive Distortion Detection, Large Language Models, Dataset-Agnostic Evaluation, Mental Health, Automated Annotation, LLM-as-Annotator} }

\begin{document}

\maketitleabstract
\section{Introduction}
Introduced by \textbf{Aaron T. Beck} \citep{beck1963thinking} and expanded by \textbf{David Burns} \citep{burns1980}, cognitive distortions (CDs) are common patterns of biased thinking that can influence how individuals perceive and interpret reality \citep{beck2009depression}. Research shows that CDs are often associated with mental health conditions such as depression \citep{joormann2016examining}, anxiety \citep{yazici2022interpersonal}, and PTSD \citep{ouhmad2024cognitive}.
These thought patterns can often be reflected in language, making them potentially detectable through natural language analysis. With the increased availability of digital text data, especially from online forums and social media, researchers have started exploring automated CD detection \citep{shickel2020cognitive, rojasbarahona2018deeplearninglanguageunderstanding, simms2017cognitive}, which in turn could support the development of automated tools for cognitive reframing of negative thoughts or reappraising events causing negative emotions.

Automated CD detection requires annotated data, either for training classification models or evaluating the accuracy of LLM-based systems using in-context learning. However, CD annotation can be subjective, as annotators may perceive and interpret distortions differently depending on their understanding of the concept or the context provided. Prior works have reported low agreement between annotators, highlighting the difficulty in achieving consistency when annotating CDs \citep{shreevastava2021detecting,lybarger-etal-2022-identifying,ding-etal-2022-improving}. This subjectivity 
can result in annotated datasets that are noisy and potentially unreliable. Models trained on such data are forced to fit these inconsistencies, which can hinder their ability to generalize and accurately detect CDs in test data. Indeed, previous works have reported weighted F1-score in a range between 0.2--0.4 \citep{chen-etal-2023-empowering,shickel2020cognitive,lim2024erd}, suggesting that models have considerable confusion in discriminating between different CD categories.

Recent works have explored LLMs as alternatives to expert or crowd-sourced annotators in text classification tasks, typically focusing on single predictions or comparisons with human annotations \citep{gilardi2023chatgpt,heseltine2024large,aldeen2023gpt}. However, relying on a single output from LLM as the final annotation can be problematic, as it raises questions about reliability and whether the label is trustworthy or merely a result of randomness in sampling.

Therefore, we hypothesize that multiple independent LLM runs can help surface labels that are consistently assigned across runs, pointing to a reliable feature characteristic of the text. The intuition is that while individual predictions may vary, the recurrence of certain labels suggests an underlying stability, offering a form of internal agreement that could reduce some of the variability seen in human annotations, leading to more robust labels.
Building on this, we propose the following contributions:
\paragraph{Contribution 1:} We propose an LLM-based annotation framework, where we run multiple independent annotation passes on the same text using GPT-based models, and select labels that appear consistently across nearly all runs, likely representing core CDs present in the text.  
We focus on producing annotations that are internally reliable by identifying labels that appear consistently across multiple runs. Our view is that reliability is a necessary first step toward assessing validity, especially in subjective domains like mental health, where ground truth is often unattainable. Without consistent behavior, neither model evaluation nor human verification can meaningfully proceed.

We verify the effectiveness of our methodology through:
\textbf{(i) Statistical validation of annotation reliability:} Where 
LLMs, especially GPT-4, with moderate temperature settings, achieve high internal consistency (Fleiss' kappa = 0.78) across multiple runs.  
\textbf{(ii) Performance on downstream task:} 
Where models trained on our LLM-generated labels consistently outperform those trained on human-annotated labels. 
\textbf{(iii) Verification by domain experts:} 
Where results were inconclusive, with no clear preference emerging between label sources. However, a low agreement between experts highlights the inherent ambiguity and subjectivity of the task and limitations in the dataset itself.

\paragraph{Contribution 2:} Re-annotation of the dataset makes direct comparison of downstream task performance difficult across the literature.
This led us to our second contribution: a dataset-agnostic evaluation methodology. We adopt the notion of the effect size calculation (commonly used in psychology and medicine), 
with the kappa measure, which accounts for both random chance and model performance, and provides a standardized scale to compare the performances of predictive models trained on datasets with differing characteristics. 
The proposed metric result shows that models trained on LLM-annotated labels consistently achieve higher scores, indicating a greater improvement over the random baseline, as compared to models trained on human-annotated labels.

\vspace{0.5em}
\noindent
In summary, we propose an LLM-based annotation method for subjective tasks, grounded in the assumption that reliable and meaningful labels will consistently emerge across multiple independent runs.
We also introduce a data-agnostic method for evaluating models trained and tested on different datasets by adapting the concept of effect size, using the kappa measure to provide a performance metric normalized against chance-level baselines.

\section{Related Work}
Several studies have explored automated CD detection using different datasets from varying sources. However, many of these efforts suffer from low inter-annotator agreement (IAA) \citep{shreevastava2021detecting} or the lack of it \citep{simms2017cognitive, aureus2021covid, shickel2020cognitive}, which raises concerns about the reliability of the annotated labels and their utility in classification tasks.
\citet{wang-etal-2023-c2d2} report high IAA scores; however, the English version of the dataset is not publicly available yet. Similarly, studies based on patient-therapist text exchange \citep{lybarger-etal-2022-identifying, tauscher2023automated, ding-etal-2022-improving} also report low to moderate IAA scores and limited classification performance, 
but the sensitive nature of the dataset makes it publicly unavailable.

LLMs like ChatGPT have emerged as scalable alternatives to human annotators, offering consistent and cost-effective labeling across various NLP tasks \citep{he2024annollmmakinglargelanguage, li2024comparativestudyannotationquality,zhang2023llmaaamakinglargelanguage, gilardi2023chatgpt}. 
Recent studies leveraged LLMs for CD classification task through prompting frameworks. The ERD framework \citep{lim2024erd} and DoT prompting \citep{chen-etal-2023-empowering} both apply multi-step reasoning to enhance CD detection. However, as they focused only on classification task based on the same publicly available but unreliable gold-standards annotations by \citet{shreevastava2021detecting} with IAA 33.7\%, the core issue of annotation quality remains unexplored.

In contrast to these studies, our work addresses the foundational problem of label reliability by introducing an annotation schema utilizing LLMs aimed at producing more consistent and reliable ground truth annotations, which automatically results in better performance on down stream tasks.

\section{CD Annotation with LLMs}
We hypothesize that the recurrence of labels across multiple independent LLM runs reflects the model detecting stable and interpretable patterns in the input text. Rather than treating single outputs as definitive annotations, we view consistent predictions as a signal of internal model reliability, particularly valuable in subjective tasks where objective ground truth is unavailable.
In this section, we first describe our data and annotation procedure, followed by a quantitative analysis of label consistency across runs. Finally, we assess the reliability of the resulting annotations using inter-run agreement metrics. Together, these steps establish the foundation for our methodology: leveraging label stability as a proxy for annotation reliability in the absence of human consensus.

\begin{table*}[th]
\centering
\small
\renewcommand{\arraystretch}{1.0}
\begin{tabular}{p{0.3cm} p{3.3cm} p{11.0cm}}
\hline
\textbf{No.}&\textbf{Cognitive Distortion} & \textbf{Description and Example} \\
\hline

1.&Emotional Reasoning & Assuming emotions reflect reality. \textit{Example:} “I feel worthless, so I must be a failure.” \\

2.&Overgeneralization & Drawing broad conclusions from limited experiences. \textit{Example:} “I failed this interview, I’ll never get a job.” \\

3.&Mental Filter & Focusing only on negative aspects. \textit{Example:} “Everyone said my presentation was good, but one person criticized it, so it must have been terrible.” \\

4.&Should Statements & Holding rigid expectations for oneself or others. \textit{Example:} “I should always be calm and never get upset.” \\

5.&All-or-Nothing Thinking & Viewing situations in extremes. \textit{Example:} “If I’m not the best, I’m a total failure.” \\

6.&Mind Reading & Presuming negative judgments from others. \textit{Example:} “She didn’t say hi, she must think I’m annoying.” \\

7.&Fortune Telling & Predicting negative outcomes with certainty. \textit{Example:} “There’s no point in applying, I know I won’t get accepted.” \\

8.&Magnification& Exaggerating potential problems. \textit{Example:} “If I mess up this report, I’ll lose my job and never recover.” \\

9.&Personalization & Taking undue responsibility for external events. \textit{Example:} “My friend is upset, it must be something I did wrong.” \\

10.&Labeling & Defining oneself or others by single traits. \textit{Example:} “I missed a deadline, I’m so incompetent.” \\
\hline
\end{tabular}
\caption{List of Cognitive Distortions (CDs) adopted from \citet{shreevastava2021detecting}. “No Distortion” is included as an 11\textsuperscript{th} category when no CD is present.}
\label{tab:cds}
\end{table*}

\subsection{Therapist Q\&A dataset}
\label{sec:data}
In this study, we use the publicly available Therapist Q\&A dataset\footnote{https://www.kaggle.com/datasets/arnmaud/therapist-qa/data\label{fn:data}}, which was annotated by \citet{shreevastava2021detecting} with ten CDs, and
we received the annotated dataset from the authors. The dataset consists of user-written texts referred to as \textsc{user input}  and corresponding responses referred to as \textsc{response} in this study. 
They annotated 2530 such pairs with 10 CDs (as shown in Table \ref{tab:cds}), 
which are referred to as \textsc{golden labels} in the remainder of this paper. 
Each \textsc{user input} within the dataset was annotated with one dominant distortion and, optionally, a secondary distortion. 
The authors measured an IAA based on about one-third of the dataset labeled by two annotators, resulting in 33.7\% using the joint probability of agreement metric. More details about the dataset and reasoning behind its selection for this study can be seen in Appendix \ref{appendix:dataset}.

\subsection{LLM Annotation Procedure}
We start by proposing the LLM-based annotation schema that forms the basis for our proposed hypothesis.
In particular, we experimented with OpenAI's GPT-4 models via the API\footnote{https://platform.openai.com/docs/api-reference} provided through Microsoft Azure\footnote{https://azure.microsoft.com/en-us}.
We selected GPT-4 and GPT-4o LLM models, which we pair with two temperature settings 0.5 and 0.7, to control the level of randomness in the generated outputs, allowing us to observe whether consistent labels still emerge under variable conditions (\textit{Detailed LLM configurations are presented Appendix \ref{appendix:llmconfig} and temperature selection in Appendix \ref{appendix:temp}}). 
We did not perform extensive hyperparameter tuning, since our goal is to assess the reliability of LLM annotations rather than optimize model performance. 
Given the sensitive nature of the dataset, Microsoft Azure's default content filter had to be disabled to allow full processing of the dataset. Furthermore, we explore two prompts to instruct the models (\textit{Detailed prompts can be seen in Appendix \ref{Appendix:prompt}}):\\

\noindent \textbf{Multi-Label Prompt (MLP):}
This prompt annotates each \textsc{user input} with one or more CDs considered in this study, allowing an unconstrained assignment of labels.\\
\noindent \textbf{Ranked-Label Prompt (RLP):} 
Following the method by \citet{shreevastava2021detecting}, this prompt constrains the model to select only the most dominant CD present in the \textsc{user input}, with the option to add a secondary distortion if applicable.\\

To test our hypothesis that whether consistent patterns emerge across independent runs, we process each \textsc{user input} through five independent API calls per prompt type.
This is done for all configurations of models and temperatures (GPT4-0.5, GPT4-0.7, GPT4o-0.5, and GPT4o-0.7), resulting in a total of 40 annotations per \textsc{user input} across all combinations. 
An example of \texttt{user input} and its CD annotations from different configurations and prompt types can be seen in Figure \ref{fig:example_annotation_iterations}. 

\begin{figure*}[t!]
  \centering
  \scriptsize
  \begin{tcolorbox}[
    colback=gray!5, colframe=gray!40,
    boxrule=0.4pt, arc=1pt,
    boxsep=2pt, left=2pt,right=2pt,top=2pt,bottom=2pt,
    width=\textwidth
  ]
  \textbf{User Input.}\;
``Hello, I have a beautiful, smart, outgoing and amazing five year old little girl. Yesterday she came to me and said mom can you take me to the doctor. I ask her what was wrong and she replied: I hear voices in my ears but I do not see the people saying it. She says it happened during school during a reading circle. She thought someone called her stupid and let the teacher know\ldots (continue text)”
  \end{tcolorbox}

  \vspace{2pt}

  \setlength{\tabcolsep}{3pt}
  \renewcommand{\arraystretch}{1.05}
  \begin{minipage}[t]{0.49\textwidth}
    \centering
    \textbf{RLP Prompt}\\[-1pt]
    \begin{tabular}{@{}lllll@{}}
      \hline
      \textbf{Run} & \textbf{GPT4-0.5} & \textbf{GPT4-0.7} & \textbf{GPT4o-0.5} & \textbf{GPT4o-0.7} \\
      \hline
      1 & Pers. & Pers., FT & ND & ND \\
      2 & Pers. & Pers., ER & ND & ND \\
      3 & Pers. & Pers., ER & ND & ND \\
      4 & Pers. & Pers., ER & ND & ND \\
      5 & Pers. & Pers., FT & ND & ND \\
      \hline
      
    \end{tabular}

  \end{minipage}
  \hfill
  \begin{minipage}[t]{0.49\textwidth}
    \centering
    \textbf{MLP Prompt}\\[-1pt]
    \begin{tabular}{@{}lllll@{}}
      \hline
      \textbf{Run} & \textbf{GPT4-0.5} & \textbf{GPT4-0.7} & \textbf{GPT4o-0.5} & \textbf{GPT4o-0.7} \\
      \hline
      1 & Pers., FT & Pers. & ND & ND \\
      2 & Pers., FT & Pers. & ND & ND \\
      3 & Pers., ER & Pers. & ND & ND \\
      4 & Pers., ER & Pers. & Pers.  & ND \\
      5 & Pers., FT & ND & Pers.  & ND \\
      \hline
    \end{tabular}
    
  \end{minipage}

  \caption{Example illustrating CD annotations for one user input across different GPT configurations, prompt types,
and runs. (Here: Pers. = Personalization, FT = Fortune Telling, ND = No Distortion, ER = Emotional Reasoning)}
  \label{fig:example_annotation_iterations}
\end{figure*}

Despite clear instructions in prompts, we noticed several instances where LLM generated new or modified labels were not present in the predefined list of CDs. 
We reviewed these cases and grouped all such labels under a single category of `Others', resulting in 12 label classes: 10 predefined CDs, No Distortion, and Others (\textit{The list of extra labels is given in Appendix \ref{Appendix:others} Table \ref{tab:other_labels_summary}}). 

\subsection{Label Consistency}
\label{sec:consistency}
\begin{figure}[t!]
    \centering
    \includegraphics[width=0.44\textwidth]{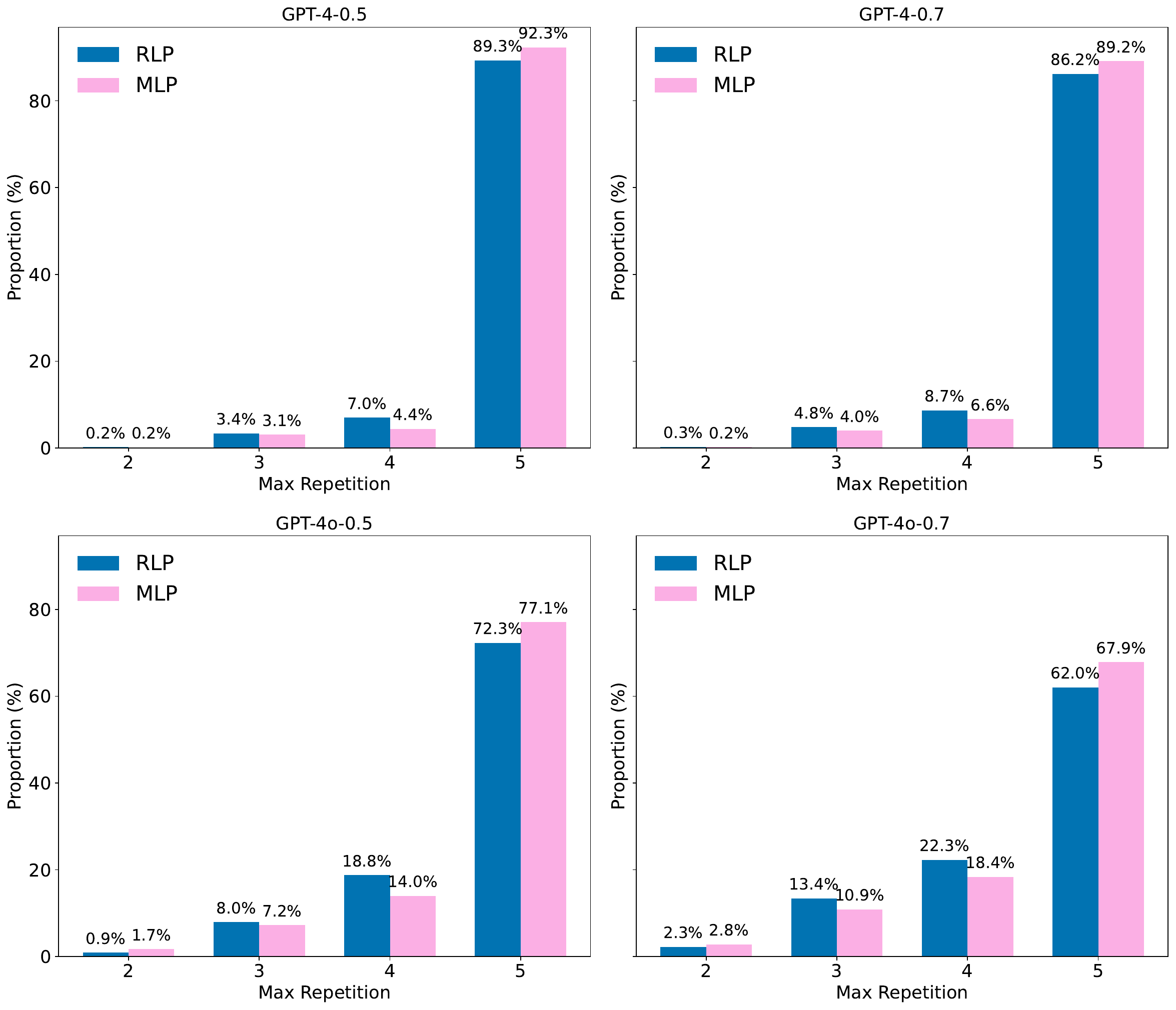}
    \caption{Distribution of maximum label repetitions across all configurations.}
    \label{fig:max_repetition_distribution}
\end{figure}
As per our hypothesis, we explore whether repeated runs of LLMs yield more stable and consistent CD labels.
To investigate this, Figure~\ref{fig:max_repetition_distribution} plots the proportion of data points having at least one label repeated a given number of times, illustrating the extent of label consistency across runs.

Plots show most data points have at least one label repeated 5 times, indicating a high model confidence in these labels. GPT-4 at 0.5 temperature shows the highest consistency, with over 90\% of data points having at least one label repeated in all runs. 
Across prompt types and temperatures, GPT-4 demonstrates greater consistency than GPT-4o, which tends to assign more varied sets of labels across runs.
Across all configurations, at least 84\% of data points have at least one label repeated four or more times.
Overall, these results provide evidence in favor of our hypothesis that repeated runs can surface labels that are consistent and are less likely to be random. 

\subsection{Reliability}
\label{sec:fleiss}
To assess the reliability of LLM-generated labels across the five independent LLM runs, we compute Fleiss' kappa \cite{fleiss1971measuring}, which is a standard statistical measure of assessing the agreement between multiple ratings of categorical values. 
Since our task involves multi-label annotations, we treat each label as a separate binary task (present vs absent), and calculate Fleiss' kappa scores per label to evaluate consistency. 
Table \ref{tab:average_fleiss_kappa_11_labels} reports Fleiss' kappa scores across configurations and prompt types averaged over the 11 CD labels considered (\textit{Fleiss kappa calculations, along with detailed scores, are
shown in Appendix \ref{appendix:fleiss_kappa} Table \ref{tab:fleiss_kappa}}).

The overall scores in Table \ref{tab:average_fleiss_kappa_11_labels} show moderate to substantial agreement between the different runs across all configurations and prompt types considered. In particular, GPT-4 model consistently achieved higher agreement scores than GPT-4o, which aligns with the consistent labeling patterns observed in Figure \ref{fig:max_repetition_distribution}. These Fleiss’ Kappa scores as inter-run agreements provide statistical evidence in support of our initial hypothesis that multiple LLM runs can surface stable and reliable labels within otherwise subjective tasks.
\begin{table}[t!]
\centering
\begin{tabular}{lcc}
\toprule
\textbf{Configurations} & \textbf{RLP} & \textbf{MLP} \\
\midrule
\textbf{GPT4-0.5} & \textbf{0.78} & \textbf{0.78} \\
\textbf{GPT4-0.7} & 0.73 & 0.71 \\
\textbf{GPT4o-0.5} & 0.63 & 0.62 \\
\textbf{GPT4o-0.7} & 0.52 & 0.54 \\
\bottomrule
\end{tabular}
\caption{Average Fleiss' kappa agreement scores across model configurations and prompt types.}
\label{tab:average_fleiss_kappa_11_labels}
\end{table}

\section{CD Annotation Validation}
In this section, we want to evaluate the practical utility of the LLM-generated labels. The motivation of this analysis stems from the concern that training machine learning models on noisy or inconsistent labels can lead to the models trying to fit to this noise, resulting in limited generalizability and poor performance on test data. Conversely, models trained on more stable and consistent labels are expected to benefit from a stronger learning signal, leading to better generalization. As such, we start by defining our final label selection process.

\subsection{Final Label Selection}
\label{sec:final_labels}
The final selection of label(s), traditionally carried out by majority voting, i.e., labels appearing in at least 3 out of 5 runs would be selected, sets a low threshold for confidence.
Instead, we prioritize labels that recur in at least 4 out of 5 runs, effectively simulating a higher confidence interval, treating it as a strong indicator of model certainty, resulting in more robust and reliable labels.
Data points that do not meet this threshold are assigned special categories to indicate annotation ambiguity (\textit{see Appendix \ref{appendix:label_dist} for more details}).

This process was repeated for all the configurations considered, providing a total of 8 sets of final labels, i.e., per configuration, we have: 1) \textsc{MLP labels}, derived from the Multi-Label Prompt, 2) \textsc{RLP labels}, derived from the Rank-Label Prompt. We also include \textsc{golden labels}, taken from \citet{shreevastava2021detecting}. The distribution of final labels can be seen in Appendix \ref{appendix:label_dist} Table \ref{tab:label distributioon}.

\subsection{Experimental Setup}

The dataset was first split into train, development, and test sets using a 70:15:15 ratio, with stratification based on the \textsc{golden labels}.
Then, we created four datasets, one for each model-temperature configuration (GPT4-0.5, GPT4-0.7, GPT4o-0.5 and GPT4o-0.7).
Consistency across datasets was ensured by keeping the split assignment fixed: a \textsc{user input} assigned to the training set, for instance, will always belong to the training set across all datasets.
After splitting, a preprocessing step was performed independently for each dataset to remove examples with ambiguous label categories (section \ref{sec:final_labels}), generated by either prompt setting. 
For example, if a given \textsc{user input} received a valid CD label from the \textsc{RLP} prompt but an ambiguous category label from the \textsc{MLP} prompt, that example was excluded from all the label settings (\textsc{RLP}, \textsc{MLP}, and \textsc{Golden labels}) within the dataset. We did this filtration to avoid introducing categories not present in the \textsc{Golden labels} for the sake of comparability. Any bias introduced as a result of this filtration would be systematic for all label types and should not hinder fair comparison. 
As a result, the number of retained examples within each split can vary slightly between datasets, resulting in four datasets of different sizes (one for each model-temperature configuration) 
as shown in Table \ref{tab: final_sets}. The train, development, and test splits maintain a 70:15:15 ratio very closely.

\begin{table}[t] 
\centering 

\begin{tabular}{lllll} 
\toprule 
\textbf{Configuration} & \textbf{Train} & \textbf{Dev} & \textbf{Test} & \textbf{Total} \\
\midrule 
Original data & 1771 & 379  & 380  & 2530\\
\midrule
GPT4-0.5                & 1667  & 346  & 356   & 2369\\ 
GPT4-0.7                & 1628  & 344   & 347   & 2319\\ 
GPT4o-0.5               & 1494 & 306   & 323   & 2123\\ 
GPT4o-0.7               & 1325 & 291   & 272   & 1888 \\ 
\bottomrule 
\end{tabular} 
\caption{Data split sizes for the original data and for the four model and temperature configurations. The splits for LLM-generated datasets are smaller than the original, because texts with ambiguous label categories have been removed.} 
\label{tab: final_sets} 
\end{table}

We use MentalRoBERTa \cite{ji2022mentalbert}  transformer-based model for three classification tasks: (1) \textbf{Binary classification} to detect the presence of CD, (2) \textbf{Multi-class classification} predicting only the dominant label of the \textsc{RLP Labels} and \textsc{Golden Labels}, and (3) \textbf{Multi-Label classification} to detect all CDs present in a data point.

We trained all models using standard fine-tuning procedures with 
five random initializations, and results were averaged. Implementation details can be found in Appendix \ref{Appendix:models}.

\subsection{Results}
\label{sec:weighted_f1}
\begin{table*}[th]
    \centering
    \small
    \setlength{\tabcolsep}{5pt}
    \renewcommand{\arraystretch}{1.05}
    \begin{tabular}{lccccccccc}
        \toprule
        \multirow{2}{*}{\textbf{Datasets}} 
        & \multicolumn{3}{c}{\textbf{RLP Labels}} 
        & \multicolumn{3}{c}{\textbf{MLP Labels}} 
        & \multicolumn{3}{c}{\textbf{Golden Labels}} \\
        \cmidrule(lr){2-4}
        \cmidrule(lr){5-7}
        \cmidrule(lr){8-10}
        & Binary & Multiclass & Multilabel 
        & Binary & Multiclass & Multilabel 
        & Binary & Multiclass & Multilabel \\
        \midrule
        Gpt4-0.5   & 0.838 & 0.559 & 0.575 & 0.831 & N/A   & \textbf{0.609} & 0.768 & 0.384 & 0.311 \\
        Gpt4-0.7   & \textbf{0.854} & \textbf{0.604} & 0.548 & 0.838 & N/A   & 0.603 & 0.770 & 0.391 & 0.332 \\
        Gpt4o-0.5  & 0.832 & 0.481 & 0.396 & 0.800 & N/A   & 0.489 & 0.778 & 0.384 & 0.287 \\
        Gpt4o-0.7  & 0.809 & 0.476 & 0.428 & 0.829 & N/A   & 0.474 & 0.813 & 0.395 & 0.338 \\
        \bottomrule
    \end{tabular}
    \caption{Weighted F1 scores (MentalRoBERTa) on corresponding test sets averaged over five initializations. N/A indicates task not applicable for the label set.}
    \label{tab:f1_weighted_scores}
\end{table*}

Table \ref{tab:f1_weighted_scores} provides the weighted F1 scores for the test sets of different datasets considered within this research, averaged over 5 random model initializations (\textit{more detailed results in appendix \ref{appendix:results}}). 
Our results show that models trained on \textsc{RLP Labels} and \textsc{MLP Labels} consistently outperform the models trained on \textsc{Golden Labels} across different classification tasks and datasets.  
This aligns with our earlier expectation that more consistent labels, such as those generated through LLM, provide a stronger training signal, allowing models to generalize more effectively.

However, it is important to emphasize that these results are only indicative and not directly comparable. First, the data splits across the four datasets are not identical. Second, even within a given dataset, although the splits remains the same, the label assignments differ across \textsc{gold labels}, \textsc{RLP labels}, and \textsc{MLP labels}.
Therefore, in the following section, we propose a methodology to compare model performances across datasets.

\section{Dataset-Agnostic Evaluation Methodology}
\label{sec:normalization}
In NLP, models and methods are typically compared using benchmark datasets. For example, the annotations provided by \citet{shreevastava2021detecting} for the Therapist Q\&A dataset, also used in this work, have served as a benchmark in prior studies \citep{chen-etal-2023-empowering,lim2024erd}. However, when the datasets have different sizes and label distributions,  
direct comparison using standard evaluation metrics becomes problematic. We faced this issue in section \ref{sec:weighted_f1}, where model performances were not directly comparable for our four datasets of different characteristics. 
In this section, we propose and implement a dataset-agnostic evaluation method for comparing model performance, based on a random baseline and the concept of effect size, which is a statistical measure commonly used to aggregate results across different datasets in fields such as psychology and medicine.

\subsection{Overview}
In general terms, effect size is a quantitative measure of the strength or magnitude of a phenomenon \citep{kelley2012effect}. Effect sizes are expressed on a standardized scale, making their magnitude relatively easy to interpret. For instance, a commonly used effect size measure, Cohen’s d, represents the difference between the means of two groups in units of standard deviation.
The advantage of using effect size is that it enables bringing various studies conducted using different configurations and datasets to the same scale and thus make them comparable.
Although effect sizes are routinely reported in psychology and medical research, the concept remains largely unexplored in NLP and machine learning (ML) more broadly. A rare exception is the paper by \citet{henderson2018distilling}, which advocates for conducting meta-analyses in ML to quantify the influence of various factors. However, the authors do not propose a concrete method for computing effect sizes.

In this paper, we propose to use the Cohen's kappa \cite{cohen1960coefficient} to quantify the effect size for predictive classification models, which accounts for both the model's performance and random chance.
Simply put, Cohen's kappa provides a normalized measure 
representative of a model's performance relative to random chance assignment. Through this approach, rather than looking at absolute performance across datasets, models are compared based on their performance improvements over the corresponding random chance assignments, providing a fair comparison across datasets with different characteristics.
The three key features to implement this method are:
\begin{enumerate}
    \item \textbf{Random baseline ($F1_{random}$)}: Weighted F1 score calculated using random label assignments as model predictions based on the dataset's observed class distributions.
    \item \textbf{Model performance ($F1_{calculated}$)}: weighted F1 scores obtained for the trained models.
    \item \textbf{Normalization ($\kappa_{F1}$)}: we then normalize the model's performance using the random baseline score with the kappa formula.
\end{enumerate}
\begin{table*}[th!]
\centering
\small
\setlength{\tabcolsep}{5pt}
    \renewcommand{\arraystretch}{1.05}
\begin{tabular}{lccccccccc}
\toprule
\multirow{2}{*}{\textbf{Datasets}} 
& \multicolumn{3}{c}{\textbf{RLP Labels}} 
& \multicolumn{3}{c}{\textbf{MLP Labels}} 
& \multicolumn{3}{c}{\textbf{Golden Labels}} \\
\cmidrule(lr){2-4}
\cmidrule(lr){5-7}
\cmidrule(lr){8-10}
& Binary & Multiclass & Multilabel 
& Binary & Multiclass & Multilabel 
& Binary & Multiclass & Multilabel \\
\midrule
Gpt4-0.5   & 0.571 & 0.438 & \textbf{0.348} & 0.574 & N/A   & 0.283 & 0.515 & 0.257 & 0.138 \\
Gpt4-0.7   & \textbf{0.572} & \textbf{0.446} & 0.331 & \textbf{0.598} & N/A   & \textbf{0.324} & 0.481 & 0.241 & 0.163 \\
Gpt4o-0.5  & 0.556 & 0.351 & 0.229 & 0.469 & N/A   & 0.317 & 0.503 & 0.254 & 0.110 \\
Gpt4o-0.7  & 0.551 & 0.309 & 0.279 & 0.531 & N/A   & 0.310 & 0.475 & 0.250 & 0.169 \\
\bottomrule
\end{tabular}
\caption{Kappa scores ($\kappa_{F1}$) for MentalRoBERTa across datasets and classification tasks on test sets. Higher values indicate greater improvement over the random baseline.}
\label{tab:kappa_calculated_scores_combined}
\end{table*}
\subsection{Random Baseline ($F1_{random}$)}
\label{sec:randomf1}
To establish a realistic performance baseline, we compute the random weighted F1 score for each dataset and classification task by assigning predictions randomly according to the class distribution observed in the corresponding ground truth labels, rather than using a uniform distribution. By aligning the predicted class probabilities with the true label distribution, this approach reflects the expected performance of a model that does not learn from the data but mirrors class imbalance. These randomly assigned labels (substitute for model predictions) and ground truth labels are then used to compute the weighted F1 score ($F1_{random}$). 

We start by providing a detailed derivation of the random F1 score used as a baseline in our proposed evaluation framework. For example, we consider a dataset with 3 classes ($A$, $B$, and $C$) with given distributions ($a$, $b$, and $c$) respectively such that $a+b+c=1$. This mean the probability of selecting label $A$ during random assignment is $P(A)=a$, with $P(B)=b$ and $P(C)=c$. This results in a confusion matrix as shown below 

\[
\begin{array}{c|ccc}
    & \text{Predicted A} & \text{Predicted B} & \text{Predicted C} \\
    \hline
    \text{True A} & a \times a\times N & a \times b\times N & a \times c\times N \\
    \text{True B} & b \times a\times N & b \times b\times N & b \times c\times N \\
    \text{True C} & c \times a\times N & c \times b\times N & c \times c\times N \\
\end{array}
\]

where $N$ is the total number of data point in the dataset and the probability of a data point belonging to class $A$ being assigned label A in random assignment is $a\times a$, giving true positive value for class A ($TP_A$) as $a \times a\times N$. The same reasoning applies to other values within the confusion matrix.

Based on this confusion matrix, precision, recall, and F1 score for class A can be calculated as:

\begin{align*}
   \text{Precision}_{A} &= \frac{TP_{A}}{TP_{A} + FP_{A}} \\
   &= \frac{a \times a \times N}{(a \times a + a \times b + a \times c) \times N} \\
   &= \frac{a^2}{a(a + b + c)} = a
\end{align*}

\begin{align*}
   \text{Recall}_{A} &= \frac{TP_{A}}{TP_{A} + FN_{A}} \\
   &= \frac{a \times a \times N}{(a \times a + a \times b + a \times c) \times N} \\
   &= \frac{a^2}{a(a + b + c)} = a
\end{align*}

\begin{align*}
    F1_A &= \frac{2 \times Precision_A \times Recall_A}{Precision_A + Recall_A} \\
    &=\frac{2 \times a \times a}{a + a} \\
    &= a
\end{align*}

Similarly, calculation for class $B$ and $C$ provide $F1_B = b$ and $F1_C = c$. 
Furthermore,

\[
    \text{Weighted F1} = a \times \text{F1}_{A} + b \times \text{F1}_{B} + c \times \text{F1}_{C} 
\]
\[
    = a^2 + b^2 + c^2    
\]

This procedure avoids sampling variance and provides a mathematical formula for a random baseline score for each dataset. To verify the accuracy and robustness of our proposed random F1 score calculation methodology, we performed empirical tests by simulating random label assignment. Empirical results closely matched the theoretical scores, with deviations well within the expected standard error, thus validating our approach (\textit{detailed results are provided in Appendix \ref{appendix:random f1}}).

\subsection{Normalization ($\kappa_{F1}$)}
The normalized effect size measure is implemented with the Cohen's kappa formula, which quantifies agreement between two sets of labels while accounting for chance, given by:
\begin{align*}
    \kappa = \frac{P_{o} - P_{e}}{1 - P_{e}}
\end{align*}

In our case, we use the weighted F1-score to quantify the agreement between the trained model and the annotated labels.
Specifically, $F1_{calculated}$ is used for the observed agreement $P_o$ and $F1_{random}$ is for the random agreement $P_e$:
\begin{align*}
    \kappa_{F1} = \frac{F1_{calculated} - F1_{random}}{1 - F1_{random}}
\end{align*}
 
This transformation allows us to express model performance on a standardized scale where 0 indicates random chance performance, 1 indicates perfect performance, and values between 0 and 1 reflect the degree to which model predictions exceed chance.
Although the level of chance performance varies across models trained on different datasets, or on the same data with different label assignments, Cohen’s kappa accounts for this by incorporating dataset-specific chance agreement, making the resulting kappa scores comparable.

\subsection{Results}
\label{sec:kappa_random_results}
Table \ref{tab:kappa_calculated_scores_combined} presents $\kappa_{\text{F1}}$ scores for our four datasets generated in the previous section across three classification tasks. These scores now enable more meaningful comparison of model performance across datasets and configurations. Consider the GPT4-0.7 configuration, a multilabel model trained on \textsc{RLP labels} provides a 33.1\% improvement, whereas models trained on \textsc{Golden labels} only manage to improve 16.1\% over the respective baselines.
Overall, models trained on LLM-generated labels (\textsc{RLP Labels} and \textsc{MLP Labels}) consistently outperform those trained on \textsc{Golden labels} across all datasets and classification tasks. This reinforces the idea that training on more consistent and reliable LLM-annotations lead to better generalization on test data, as opposed to training on noisy or inconsistent labels that hinder model performance.

\begin{figure*}[h]
    \centering
    \includegraphics[width=0.9\textwidth]{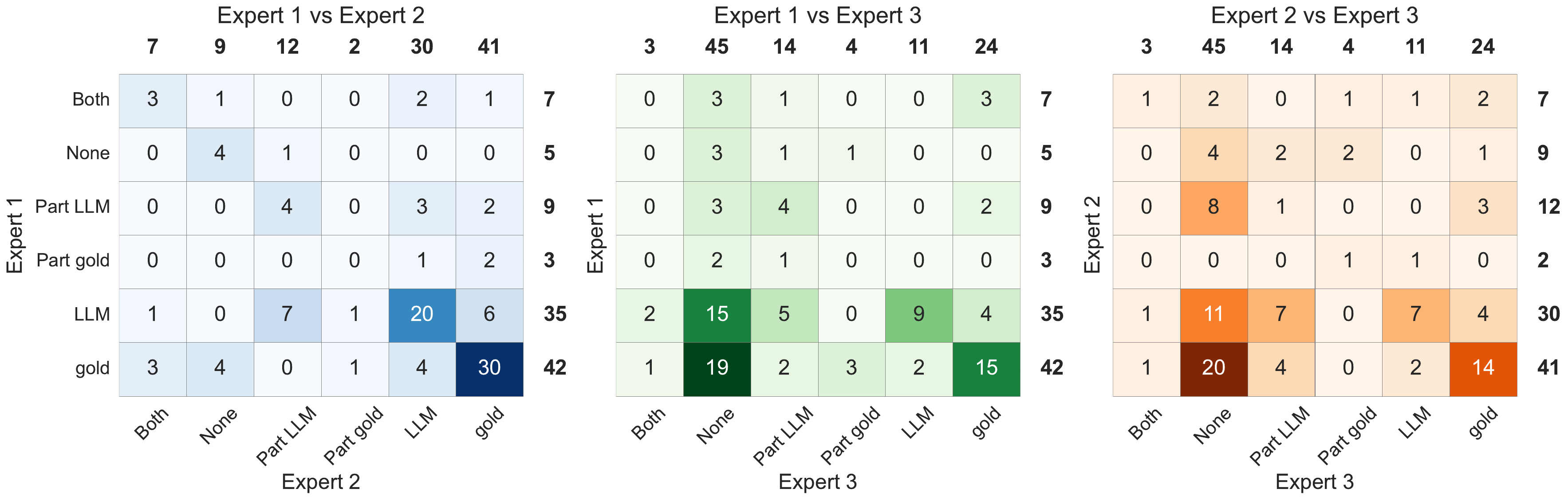}
    \caption{
    Pairwise confusion matrices b/w experts. (\textit{Here, gold = \textsc{golden labels}, LLM = LLM-generated labels, part gold = Partial \textsc{golden labels}, part LLM = Partial LLM-generated labels.}) Values on the right and top of the plots represent the total number of data points within each category for respective experts.}
    \label{fig:expert_confusion_matrices}
\end{figure*}

\section{Human Verification}
While LLM-generated labels showed higher consistency and downstream model performance, it remains essential to assess the quality of these labels through human oversight. To this end, we conducted an expert verification experiment to validate the quality of our LLM-generated labels compared to \textsc{golden labels}.

For this process, 101 samples were randomly chosen from subset of the GPT4-0.5 dataset where LLM-generated labels and \textsc{golden labels} disagreed. GPT4-0.5 dataset was chosen based on its higher inter-run agreement 
(Table \ref{tab:average_fleiss_kappa_11_labels}) and predictive model performance 
(Table \ref{tab:kappa_calculated_scores_combined}).
Three psychology experts\footnote{Two master-level practicing clinicians (one of them is also a co-author), and a Phd-level psychology researcher. All have at least introductory level training in CBT.} were asked to review the samples through Label Studio\footnote{https://labelstud.io/} annotation tool. 
They were shown the \textsc{user input} and a randomized, anonymized pair of LLM-generated and \textsc{golden labels} (Label1 and Label2), and asked which they agreed with: (1)Label1, (2)Label2, (3)Both, (4)None, (5)Partially Label1, or (6)Partially Label2.
The experts were given the list of CDs together with their description as given by \citet{burns1989feeling} as annotation guidelines (\textit{more details in Appendix \ref{appendix:sampling}}).

Experts' label selection and corresponding agreements can be seen in 
Figure \ref{fig:expert_confusion_matrices}. 
We observe that all three experts were nearly evenly split in their choices between the LLM-generated and Golden labels (including partial agreement), suggesting no strong preference for either label category. However, Expert 3 selected `None' in 45\% of the cases, indicating high disagreement with both label categories, in contrast to Experts 1 and 2, with 5\% and 9\% disagreement, respectively.

The inter-annotator agreement (Fleiss’ kappa) among the three experts yielded a low score of 0.20. Furthermore, the pair-wise agreement
between Expert 1 and Expert 2 was moderate ($\kappa$ = 0.44), while agreement between Expert 1 and Expert 3 ($\kappa$ = 0.16) and between Expert 2 and Expert 3 ($\kappa$ = 0.11) were notably lower, reflecting greater disagreement.
Thus, we conclude this verification process could not prove or disprove the validity of either LLM-generated or human-annotated labels, with expert opinions evenly split between the two.

\begin{table*}[th!]
\centering
\small
\setlength{\tabcolsep}{2pt}
\renewcommand{\arraystretch}{0.9}
\begin{tabularx}{\textwidth}{
>{\raggedright\arraybackslash}X|
>{\raggedright\arraybackslash}p{0.9cm}|
>{\raggedright\arraybackslash}p{0.9cm}
}
\toprule
\textbf{User Text}  & \textbf{Gold} & \textbf{LLM} \\ 
\midrule
I am writing because my boyfriend and I have a lot of problems in the one year we've been together. Six months ago we went on a break because I wanted to live with him but he didn't want to live with me. Even though I didn't want to end it, the arguments we had over the living together issue seemed to push him to the point of wanting to leave.
 & MR & ND\\ 
\midrule
From a young woman in Bangladesh: I have been in a very physically and mentally abusive marriage for 4 years now. I tried my best to make my marriage work and meet up to my husband's and his family's expectations, but I am always being told that I am good for nothing and I should probably kill myself. I have been accused of infidelity multiple times even when I had never done anything like that. But recently, I just couldn't tolerate all that anymore.
 & Over & ER, AoN \\ 

\bottomrule
\end{tabularx}
\caption{Example of \textsc{User inputs} in data. (\textit{Here: Gold = \textsc{Golden labels}, LLM = LLM-generated labels, MR = Mind Reading, ND = No Distortion, Over = Overgeneralization, AoN = All or nothing Thinking, ER = Emotional Reasoning}) }
\label{tab:annotation_comparison}
\end{table*}

\section{Discussion}
Given the subjectivity of the task, this study first focuses on the fundamental issue: can we produce annotations that are both internally reliable and reproducible? Our proposed method, based on repeated LLM sampling and consistency filtering, offers a systematic way to reduce annotation noise and enhance reproducibility. 
For tasks where ground truth is inaccessible and IAA (agreement between experts) is low, establishing reliability is not just helpful but essential for enabling any future progress on assessing or improving label quality. We further argue that reliability is a necessary first step towards assessing validity, especially in the mental health domain, where ground truth is often unattainable. Without consistent behavior, neither model evaluation nor human verification can meaningfully proceed.

To that end, we explored the use of LLMs as annotators, not in the conventional one-off prediction setting, where a single prediction is taken as final (may reflect stochastic variation rather than the model's true confidence in a label), but as a consistent pattern detector through repeated querying across multiple independent runs. Our hypothesis was that multiple LLM runs can help surface labels that are consistently assigned across runs, pointing to features that are reliable characteristics of the text. Experimental results show that LLM-generated labels have high inter-run agreement (Fleiss’ Kappa = 0.78 Section \ref{sec:fleiss}) as well as higher downstream task performance compared to \textsc{golden labels} (weighted F1 Section \ref{sec:weighted_f1}, and $\kappa_{\text{F1}}$ scores Section \ref{sec:kappa_random_results}). 
This gain in performance likely stems from the greater consistency of our LLM-generated labels, which offer a clearer training signal. In contrast, inconsistent labels introduce noise, hindering learning and reducing generalization, an issue seen with \textsc{golden labels}. These results not only support the use of LLMs as reliable annotators for generating consistent CD labels but also highlight the need for multiple LLM runs rather than relying on a single prediction. 

In the second part of this study, we focus on the comparability problem, how to meaningfully evaluate models when datasets differ in size, label distribution, or annotation quality. Traditional evaluation procedures used in NLP and ML rely on comparing evaluation measures obtained on the same benchmark dataset. However, if the dataset composition or label assignment differs, the absolute values of the evaluation measures are no longer comparable. To overcome this limitation, we propose a kappa-based effect size measure ($\kappa_{\text{F1}}$) that normalizes model performance relative to chance, enabling dataset-agnostic and cross-study comparison. Crucially, this approach is not limited to CD detection: it generalizes to any domain where standardized baselines are unavailable or unreliable and where datasets vary in composition. Such a measure provides a principled and interpretable foundation for comparing model performance across heterogeneous datasets, promoting fairness and reproducibility in evaluation.

Although the LLM-generated labels proved to be more consistent and less noisy than the initial \textsc{golden labels}, human evaluation involving domain experts proved to be inconclusive.
Upon examining examples (Table \ref{tab:annotation_comparison}) where experts disagreed, we observed that many of these \textsc{user inputs} do not explicitly state thoughts, but instead describe events, emotions, or experiences. While distorted thought patterns may underlie such accounts, the descriptions alone are often insufficient to reliably identify specific cognitive distortions. In a clinical setting, a therapist would typically first probe further into the user’s underlying thoughts before identifying a specific distortion. This points towards the data source itself being a limiting factor, and we therefore suggest the development of methodologies that ensure that distorted thought patterns are adequately expressed within the text.

\section{Conclusion}
This study demonstrates that Large Language Models can serve as consistent and reliable annotators for inherently subjective tasks such as cognitive distortion detection. By reframing annotation as a process of repeated querying rather than single-shot prediction, we show that multiple independent LLM runs can reveal stable labeling patterns that improve inter-run agreement and yield superior downstream model performance. These findings highlight the potential of LLMs to generate internally coherent annotations that reduce noise and enhance reproducibility, an essential step toward establishing reliability in domains where ground truth is uncertain.
Beyond annotation reliability, we addressed the broader challenge of model comparability across datasets with differing characteristics. The proposed kappa-based effect size measure ($\kappa_{\text{F1}}$) offers a dataset-agnostic evaluation framework that normalizes model performance relative to chance, enabling fairer and more interpretable comparisons across heterogeneous datasets and studies. Importantly, this framework generalizes beyond NLP to any field where benchmarks are unstable or direct comparison is not feasible. Finally, low agreement observed in human validation suggests that data lacking sufficiently articulated thought content may be inadequate for CD detection, calling for either more carefully curated datasets or methods that first identify and elicit the necessary cognitive elaboration before prediction.

\section*{Limitations}
While our study makes useful contributions, it also has limitations.
First, the annotation process relied exclusively on models from the GPT family. While these models demonstrated strong performance and consistency across runs, we did not explore the applicability or reliability of other proprietary large language models, such as Claude, Gemini, or open-source models, such as LLaMA or Mistral. As such, it remains unclear whether similar annotation quality and agreement can be achieved using non-OpenAI LLMs. We believe this represents a promising future direction for our work, involving a systematic comparison of annotation consistency and quality across different LLM families.

Second, although our methodology achieved consistency and reliability on the Therapist Q\&A dataset, several dataset-specific limitations must be acknowledged. We observed that many of the \textsc{user inputs} do not explicitly express distorted thoughts but instead describe events, emotions, or experiences. While distorted thought patterns may underlie such accounts, the descriptions alone are often insufficient to reliably identify specific distortions. In a clinical context, a therapist would typically probe further to uncover the client’s underlying distortions before assigning a label. This indicates that the data source itself imposes a structural limitation: it captures surface descriptions rather than explicit cognitive patterns, which constrains interpretability and diagnostic precision.
Furthermore, the dataset’s source cannot be independently verified, which also raises questions about authenticity and representativeness. These factors suggest that, while our framework is methodologically sound, the dataset may not be ideal for clinically oriented or context-dependent applications. Nevertheless, this dataset was selected because it remains one of the few publicly available and comparatively higher-quality resources for cognitive distortion detection and is already in active use within the research community. Its limitations underscore a broader issue within the domain, i.e., the urgent need for contextually richer datasets that more directly capture distorted thought patterns in text.

\section*{Acknowledgments}
This research was supported by the Estonian Research Council Grant PSG721 and by the Estonian Centre of Excellence in AI (EXAI). We thank the two annotators Pärtel Poopuu and Helen Uusberg.

\section*{Ethics Statement}
This study makes exclusive use of the publicly available Therapist Q\&A dataset and open-source language models. The dataset contains anonymized text that does not include any personally identifiable information. All analyses were conducted in accordance with the dataset’s terms of use. No human subjects were directly involved, and no additional data were collected, ensuring compliance with ethical standards for secondary data research.  

The broader goal of this work is to improve the reliability and transparency of automated annotation methods for subjective psychological constructs. While our findings may support the development of tools for mental health research, they are not intended for clinical diagnosis or therapeutic decision-making. We encourage responsible use of these methods within appropriate research and ethical boundaries.

\section*{Data and Code Availability}
The dataset used in this study is publicly available on Kaggle\footref{fn:data}. The code and LLM-generated labels developed for this work can be seen at GitHub\footnote{\url{https://github.com/nehasharma666/llm-cognitive-distortion-detection}}
.
\section*{References}
\bibliographystyle{lrec2026-natbib}
\bibliography{lrec2026-example}

\section*{Appendix}
\appendix
\section{Therapist Q\&A dataset}
\label{appendix:dataset}
This section contains our reasoning behind using this dataset. When we began this study, this was the most suitable publicly available resource for cognitive distortion detection.  
The dataset has been widely used and cited in the field, including in prior works such as \cite{shreevastava2021detecting, chen-etal-2023-empowering,lim2024erd}. Other available datasets in this domain \cite{simms2017cognitive,aureus2021covid,shickel2020cognitive} are, in our assessment, lower in quality, either due to annotation inconsistency, size, or lack of contextual richness. There are also non-public datasets from studies such as \cite{lybarger-etal-2022-identifying,tauscher2023automated,ding-etal-2022-improving} which we could not access due to ethical and privacy constraints. \citet{wang-etal-2023-c2d2} produced a Chinese-language dataset, but no English version is currently available. We analyzed a small sample from the Chinese version and found that the text fragments were too short and contextually sparse to reliably detect cognitive distortions. 
Unfortunately, the source of the Therapist Q\&A dataset could not be verified; moreover, the identity of the responders is also not confirmed. However this is irrelevant to our work, as our analysis focuses solely on the \textsc{user input}.

Given these limitations, we made the pragmatic decision to use the dataset in question because it was: publicly available, of comparatively higher quality than other open datasets, and already in active use by the research community. While we cannot verify the source of the data, our analyses suggest that the language in the dataset contains meaningful cognitive distortion signals to some extent and serves as a reasonable testbed for developing and evaluating annotation methodologies.

\section{Large Language Models}
\label{appendix:llmconfig}
Both LLM models considered in this research are detailed below:
\begin{itemize}
    \item \textbf{GPT-4}: Version 0613, capable of processing up to 110K tokens/minute, with a rate limit of 60 requests/minute. 
    \item \textbf{GPT-4o}: Version 2024-05-13, designed to handle up to 30M tokens/minute and up to 180K requests/minute. 
\end{itemize}

\section{Temperature}
\label{appendix:temp}
This section explains our choice of temperatures 0.5 and 0.7. We initially began our annotation process using GPT-4 with a temperature of 0.7, as it is the default setting. When switching to GPT-4o, we noticed that outputs at 0.7 were more varied across runs compared to GPT-4 at the same setting, resulting in noticeably lower consistency. 
Our choice of temperature 0.5 and 0.7 was based on trial and error. We want to observe: enough diversity to reflect the model’s interpretive range on subjective input, while still allowing for patterns of consistency to emerge where the model repeatedly selects the same label across runs. The goal is not to maximize agreement artificially, but to explore whether the model could identify stable signals in the text under mild randomness.
We intentionally avoided using a fully deterministic setting (e.g., temperature = 0), because evaluating consistency in such a deterministic system would not be logical as we can no longer evaluate whether a label appears consistently because it is strongly indicated by the input, or simply because the decoding process collapses to a single option. Such results would be trivially consistent, but not informative about the model's internal uncertainty or label robustness. Evaluation of the consistency across runs is meaningful only when the model is allowed some extent of variability in its output.

\section{Annotation Prompts} 
\label{Appendix:prompt}

\paragraph{Ranked-Label Prompt (RLP):} 
As a specialized annotator in psychology with expertise in cognitive distortions, analyze the \texttt{user input} to identify any underlying cognitive distortion(s) from the specified list called \textsc{Cognitive Distortions List}. Your task is to determine the most dominant cognitive distortion present. If there is a secondary distortion, note it as well. In cases where multiple distortions are present, select the most dominant one as the primary label, and if necessary, include one secondary label. The response must include at most two labels. If no distortions are found, label the \texttt{user input} as \textbf{No Distortion}.

\textsc{Cognitive Distortions List}:  [Emotional Reasoning, Overgeneralization, Mental Filter, Should Statements, All or Nothing Thinking, Mind Reading, Fortune Telling, Magnification (Catastrophizing), Personalization, Labeling]

\textbf{Output Format}:  If cognitive distortions are identified, provide them as a comma-separated list, with the most dominant distortion listed first, followed by a secondary distortion if applicable. At most two distortions should be listed. If no distortions are found, return \textbf{No Distortion}.

\begin{table*}[t!]
    \centering
    \small
    \renewcommand{\arraystretch}{1.3}
    \setlength{\tabcolsep}{8pt}
    \begin{tabular}{c|p{6cm}|p{6cm}}
        \hline
        \textbf{Config} & \textbf{MLP (Other Labels)} & \textbf{RLP (Other Labels)} \\
        \hline
        GPT4-0.5 & 
        Lack of Emotional Empathy, Paranoid Thinking, Procrastination, Paranoia, Delusions, Minimization, Hallucinations, Lack of Empathy, Anxiety. \newline \textbf{Total: 46} & 
        Paranoia, Fantasy, Social Avoidance, Lack of Empathy, Obsession, Comparison, Minimization, Blaming, Paranoid Thinking, Guilt, Procrastination, Rumination, Social Anxiety, Secret Keeping, Delusions, Paranoia (Mind Reading), ADHD, Low Self-Esteem, Delusional Disorder, Avoidance, Trust Issues, Dependent Personality Disorder, Delusional Thinking, Unhealthy Lifestyle. \newline \textbf{Total: 221} \\
        \hline
        GPT4-0.7 & 
        Avoidance, Comparison, Delusions, Delusions of Grandeur, Delusions of Persecution, Delusions of Reference, Hallucinations, Lack of Emotional Empathy, Lack of Empathy, Minimization, Paranoia, Paranoid Thinking, Procrastination, Social Comparison \newline \textbf{Total: 53} & 
        Avoidance, Blaming, Comparison, Delusion, Delusional Disorder, Delusional Thinking, Dependence, Disregard for Others' Feelings, Fantasy, Fantasy Thinking, Guilt, Hallucinations, Jumping to Conclusions, Lack of Empathy, Narcissism, Obsession, Paranoia, Paranoia (Mind Reading), Paranoid Thinking, Procrastination, Rumination, Self-harm, Social Anxiety, Social Avoidance \newline \textbf{Total: 207} \\
        \hline
        GPT4o-0.5 &Guilt Tripping, Rumination, Trust Issues \newline \textbf{Total: 4}   & -----\\
        \hline
        GPT4o-0.7 & Blaming, Guilt Tripping, Rumination, Trust Issues \newline \textbf{Total: 6} & Minimization, Rumination \newline \textbf{Total: 2}  \\
        \hline
    \end{tabular}
    \caption{List of labels categorized as `Others' for each model-temperature configuration and prompt type, along with total occurrences.}
    \label{tab:other_labels_summary}
\end{table*}
\paragraph{Multi-Label Prompt (MLP):} 
As a specialized annotator in psychology with expertise in cognitive distortions, analyze the \texttt{user input} to identify any underlying cognitive distortion(s) from the specified list called \textsc{Cognitive Distortions List} and provide results as per the \textbf{Output Format} only.

\textsc{Cognitive Distortions List}:  [Emotional Reasoning, Overgeneralization, Mental Filter, Should Statements, All or Nothing Thinking, Mind Reading, Fortune Telling, Magnification (Catastrophizing), Personalization, Labeling]

\textbf{Output Format}:  If any cognitive distortion(s) are identified, list them as a comma-separated list. If no distortion(s) are found, return \textbf{No Distortion}.

\section{Extra Labels} 
\label{Appendix:others}
Table \ref{tab:other_labels_summary} presents the extra labels annotated by LLMs.

\section{Fleiss' Kappa}
\label{appendix:fleiss_kappa}

\begin{table*}[t!]
\small
\centering
\begin{tabular}{lcccccccc}
\toprule
\textbf{Cognitive Distortion} & \multicolumn{2}{c}{\textbf{GPT4-0.5}} & \multicolumn{2}{c}{\textbf{GPT4-0.7}} & \multicolumn{2}{c}{\textbf{GPT4o-0.5}} & \multicolumn{2}{c}{\textbf{GPT4o-0.7}} \\
\cmidrule(lr){2-3}
\cmidrule(lr){4-5}
\cmidrule(lr){6-7}
\cmidrule(lr){8-9}
 & \textbf{RLP} & \textbf{MLP} & \textbf{RLP} & \textbf{MLP} & \textbf{RLP} & \textbf{MLP} & \textbf{RLP} & \textbf{MLP} \\
\midrule
All or Nothing Thinking & 0.66 & 0.78 & 0.60 & 0.72 & 0.48 & 0.62 & 0.40 & 0.55 \\
Emotional Reasoning & 0.81 & 0.84 & 0.76 & 0.79 & 0.56 & 0.49 & 0.44 & 0.43 \\
Fortune Telling & 0.80 & 0.834 & 0.73 & 0.78 & 0.59 & 0.65 & 0.50 & 0.58 \\
Labeling & 0.87 & 0.76 & 0.81 & 0.72 & 0.66 & 0.64 & 0.56 & 0.55 \\
Magnification & 0.82 & 0.78 & 0.75 & 0.72 & 0.63 & 0.66 & 0.53 & 0.59 \\
Mental Filter & 0.46 & 0.55 & 0.42 & 0.40 & 0.40 & 0.26 & 0.16 & 0.14 \\
Mind Reading & 0.84 & 0.79 & 0.80 & 0.73 & 0.73 & 0.68 & 0.62 & 0.59 \\
Overgeneralization & 0.73 & 0.74 & 0.68 & 0.65 & 0.58 & 0.62 & 0.48 & 0.51 \\
Personalization & 0.81 & 0.81 & 0.76 & 0.77 & 0.64 & 0.63 & 0.54 & 0.54 \\
Should Statements & 0.82 & 0.74 & 0.78 & 0.68 & 0.75 & 0.69 & 0.63 & 0.59 \\
No Distortion & 0.93 & 0.92 & 0.91 & 0.90 & 0.92 & 0.89 & 0.90 & 0.86 \\

\midrule
\textbf{Average 
} & 0.78 & 0.78 & 0.73 & 0.71 & 0.63 & 0.62 & 0.52 & 0.54 \\

\bottomrule
\end{tabular}
\caption{Fleiss' kappa agreement scores for ten cognitive distortion labels plus the No Distortion label across different GPT-4 models and temperature settings.}
\label{tab:fleiss_kappa}
\end{table*}

Fleiss' kappa is a statistical measure used to evaluate the consistency of agreement among multiple raters when they classify or categorize items. Like other kappa statistics (e.g., Cohen’s kappa), it accounts for the level of agreement expected by chance. While Cohen’s kappa is designed for two raters, Fleiss’ kappa generalizes the approach to settings with more than two raters, making it well-suited for measuring inter-run agreement across multiple LLM outputs in our study.

\vspace{0.2cm}
\noindent \textbf{Steps to Compute Fleiss' Kappa:}

\noindent \textbf{Step 1: Binary Label Conversion:} Each label is represented in a binary format, where:

\[
x_{ij} = 
\begin{cases} 
1 & \text{if label } j \text{ is present in iteration } i, \\
0 & \text{if label } j \text{ is absent in iteration } i.
\end{cases}
\]

This creates a binary matrix \( X \) of size \( N \times M \), where \( N \) is the number of items and \( M \) is the number of iterations.

\paragraph{Step 2: Construct Contingency Tables:}

For each item \( i \), calculate the number of agreements (\( n_{ij} \)) for presence (\( 1 \)) and absence (\( 0 \)) across all iterations:

\[
n_{i1} = \sum_{j=1}^{M} x_{ij}, \quad n_{i0} = M - n_{i1}.
\]

This can be visualized as a contingency table:

\[
\begin{array}{|c|c|c|}
\hline
\text{Item} & n_{i1} \, (\text{Presence}) & n_{i0} \, (\text{Absence}) \\
\hline
1 & n_{11} & n_{10} \\
2 & n_{21} & n_{20} \\
\vdots & \vdots & \vdots \\
N & n_{N1} & n_{N0} \\
\hline
\end{array}
\]

\noindent \textbf{Step 3: Compute Fleiss' Kappa:}

A. \textbf{Proportion of Agreement for Each Item}:

\[
P_i = \frac{n_{i1}(n_{i1} - 1) + n_{i0}(n_{i0} - 1)}{M(M - 1)}.
\]

B. \textbf{Overall Agreement}:

\[
P = \frac{1}{N} \sum_{i=1}^{N} P_i.
\]

C. \textbf{Expected Agreement}:

\[
P_e = \sum_{k=1}^{2} \left( \frac{\sum_{i=1}^{N} n_{ik}}{N \cdot M} \right)^2.
\]

D. \textbf{Fleiss' Kappa}:

\[
\kappa = \frac{P - P_e}{1 - P_e}.
\]

Computed Fleiss' Kappa scores for 11 CDs across different configurations and prompt type can be seen in Table \ref{tab:fleiss_kappa}.

\section{Final Label Selection}
\label{appendix:label_dist}

As explained in Section \ref{sec:final_labels}, we selected only those labels that appear at least four times across five runs. Data points that fail to meet this condition are considered too ambiguous to be consistently labeled by LLMs. 
For \textsc{user inputs} where no label reached the threshold of four recurrences, we introduced two fallback categories to reflect the nature of the ambiguity. If none of the labels met the threshold but `No Distortion' appeared in at least one run, we labeled the instance as `not sure if distortion,' acknowledging the uncertainty around whether the text contains a distortion at all. In contrast, if the labels were too diverse without reaching any consensus and `No Distortion' was absent, we marked it as `not sure which distortion,' indicating the presence of some distortions but not enough agreement to identify it. 

Before selecting the final labels, we addressed the presence of additional labels categorized as `Others' in our dataset. We notice that most of the time in runs, `Others' was accompanied by at least one CD label from the given list. In such cases, it was simply ignored and further processing was done based on the remaining labels.
The `Others' label was retained only in cases where it was not accompanied by any other CD label.

Table \ref{tab:label distributioon} presents the distribution of final cognitive distortion labels obtained from both the Ranked-Label Prompt (RLP) and Multi-Label Prompt (MLP) across four model configurations (GPT-4 and GPT-4o at temperatures 0.5 and 0.7). 
For each configuration, the table reports both the absolute number of times each label was assigned and the corresponding proportion in percentage in the dataset. As this is a multilabel annotation setting, the total percentages across labels exceed 100\%, since a single input can be assigned multiple distortion labels. The variation in label frequency across configurations highlights the influence of model type, temperature setting, and prompt strategy on annotation behavior.

\begin{table*}[t!]
\small
\centering
\resizebox{\textwidth}{!}{%
\begin{tabular}{lllllllll}
\toprule
\textbf{Cognitive Distortion} & \multicolumn{2}{c}{\textbf{GPT-4 0.5}} & \multicolumn{2}{c}{\textbf{GPT-4 0.7}} & \multicolumn{2}{c}{\textbf{GPT-4o 0.5}} & \multicolumn{2}{c}{\textbf{GPT-4o 0.7}} \\
\cmidrule(lr){2-3}
\cmidrule(lr){4-5}
\cmidrule(lr){6-7}
\cmidrule(lr){8-9}
 & \textbf{RLP } & \textbf{MLP} & \textbf{RLP} & \textbf{MLP} & \textbf{RLP} & \textbf{MLP} & \textbf{RLP} & \textbf{MLP} \\
\midrule
All or Nothing Thinking & 103 (4.1\%) &380 (15.0\%) &86 (3.4\%)	&331 (13.1\%)&34 (1.3\%)	&247 (9.8\%)&33 (1.3\%)	&213 (8.4\%)\\
Emotional Reasoning & 959 (37.9\%)&	947 (37.4\%)&932 (36.8\%)&	886 (35.0\%)&296 (11.7\%)	&119 (4.7\%)&202 (8.0\%)	&110 (4.4\%)\\
Fortune Telling & 218 (8.6\%)	&555 (21.9\%)&202 (8.0\%)	&507 (20.0\%)&53 (2.1\%)	&137 (5.4\%)&54 (2.1\%)	&131 (5.2\%)\\
Labeling & 53 (2.1\%)	&73 (2.9\%)&49 (1.9\%)	&68 (2.7\%)&347 (13.7\%)	&275 (10.8\%)&305 (12.1\%)&	244 (9.6\%)\\
Magnification & 274 (10.8\%)&	298 (11.8\%)&234 (9.3\%)	&252 (10.0\%)&481 (19.0\%)	&570 (22.5\%)&437 (17.3\%)	&532 (21.0\%)\\
Mental Filter & 8 (0.3\%)&	31 (1.2\%)&12 (0.5\%)&	22 (0.9\%)&9 (0.4\%)&	4 (0.2\%)&4 (0.2\%)	&3 (0.1\%)\\
Mind Reading & 211 (8.3\%)	&312 (12.3\%)&187 (7.4\%)	&292 (11.5\%)&141 (5.6\%)&	175 (6.9\%)&127 (5.0\%)	&165 (6.5\%)\\
Overgeneralization & 159 (6.3\%)	&217 (8.6\%)&152 (6.0\%)	&195 (7.7\%)&410 (16.2\%)	&382 (15.1\%)&362 (14.3\%)	&301 (11.9\%)\\
Personalization &656 (25.9\%)	&713 (28.2\%)&597 (23.6\%)	&676 (26.7\%)&250 (9.9\%)&	338 (13.4\%)&201 (7.9\%)&	301 (11.9\%)\\
Should Statements & 45 (1.8\%)	&80 (3.2\%)&39 (1.5\%)	&74 (2.9\%)&28 (1.1\%)	&61 (2.4\%)&24 (1.0\%)	&54 (2.1\%)\\
Not sure if distortion &	64 (2.5\%)&	81 (3.2\%)&99 (3.9\%)&	105 (4.2\%)&83 (3.3\%)	&138 (5.5\%)&113 (4.5\%)	&183 (7.2\%)\\
Not sure which distortion &28 (1.1\%)	&1 (0.04\%)&30 (1.2\%)	&1 (0.0\%)&143 (5.7\%)	&88 (3.5\%)&284 (11.2\%)&	165 (6.5\%)\\
Others &	2 (0.1\%)&	0 (0.0\%)&2 (0.1\%)	&0 (0.0\%)&	0 (0.0\%) 	&0 (0.0\%) &0 (0.0\%) 	&0 (0.0\%) \\
No Distortion & 708 (28.0\%)	&812 (32.1\%)&707 (27.9\%)&	792 (31.3\%)&665 (26.3\%)	&717 (28.3\%)&636 (25.1\%)	&688 (27.2\%)\\
\bottomrule
\end{tabular}
}
\caption{Distribution of final labels from both prompts across four model configurations. Values show the number of occurrences of CDs as well as the corresponding proportions.}
\label{tab:label distributioon}
\end{table*}

\section{Cognitive Distortion Modeling}
\label{Appendix:models}
\paragraph{Total Model Configurations and Training Experiments:} For a comprehensive evaluation, we initially used the transformer-based model MentalRoBERTa, as it has shown strong performance on mental health-related tasks. We also trained MentalBERT for comparison purposes. However, since the performance trends across both models were consistent, we included only MentalRoBERTa results in the main text for clarity. The training details are detailed below:

\begin{itemize}
    \item \textbf{Two models:} MentalBERT and MentalRoBERTa  
    \item \textbf{Four configurations:} GPT-4 (T=0.5), GPT-4 (T=0.7), GPT-4o (T=0.5), GPT-4o (T=0.7)
    \item \textbf{Three label sets:} \textsc{RLP Labels}, \textsc{MLP Labels}, and \textsc{Golden Labels}
    \item \textbf{Three classification tasks:}
        \begin{itemize}
            \item \textbf{Binary classification:} In this approach, labels were classified as binary (1 or 0), where 1 indicated the presence of any cognitive distortion, and 0 represented "No Distortion." 
            \item \textbf{Multi-class classification:} Only the dominating \textsc{RLP Labels} and \textsc{Golden Labels} were used for this approach, where dominating labels refer to the first label in the case of a multi-label scenario.
            \item \textbf{Multi-label classification:}  The multilabel binarizer was applied to each label column, where each instance could have multiple cognitive distortion labels assigned.
        \end{itemize}
\end{itemize}
For each configuration considered, the appropriate model was loaded through Hugging Face's Transformer library\footnote{https://huggingface.co/} based model-specific paths, ensuring consistent use of pre-trained parameters. The training was run for maximum of 100 epochs, with early stopping based on F1 score performance. Random initialization variability is taken into account by training for 5 different seed values, with the results averaged across seed values to report mean scores and standard deviations. Overall, the model performance is analyzed across different label sets, classification objectives, and model architectures, providing robust experimental results.
Given these factors, the total number of trained models are:

\begin{equation}
\begin{aligned}
    & 2 \text{ (M)} \times 4 \text{ (D)} \times 
    3 \text{ (L)} \times 1 \text{ (Binary)} \\
    & + 2 \text{ (M)} \times 4\text{ (D)} \times 2 \text{ (L)} \times 1 \text{ (Multiclass)} \\
    & + 2 \text{ (M)} \times 4\text{ (D)} \times 3 \text{ (L)} \times 1 \text{ (Multilabel)} \\
    &=24+16+24 = 64  \text{Models}
\end{aligned}
\end{equation}
\textbf{Where:}
M = models, D = configurations based on GPT models and temperatures, L = Label sets.

Since we trained each model using \textbf{five different seeds} for robust evaluation, the total number of training runs amounts to:
\begin{equation}
    64 \text{ models} \times 5 \text{ seeds} = 320 \text{ trained models}
\end{equation}

This setup ensures that our findings are statistically robust and account for variability due to random initialization.
\begin{table*}[t!]
    \centering
    \small
    \begin{tabular}{llccc}
        \hline
        \textbf{Labels} & \textbf{Datasets} & \textbf{Binary} & \textbf{Multiclass} & \textbf{Multilabel} \\
        \hline
        \multicolumn{5}{l}{\textbf{MentalRoBERTa}} \\
         \toprule
        \multirow{4}{*}{RLP Labels} 
            & Gpt4-0.5 & 0.838±0.006 & 0.559±0.011 & 0.575±0.012 \\
            & Gpt4-0.7 & \textbf{0.854±0.007} & \textbf{0.604±0.030} & 0.548±0.014 \\
            & Gpt4o-0.5 & 0.832±0.012 & 0.481±0.017 & 0.396±0.095 \\
            & Gpt4o-0.7 & 0.809±0.017 & 0.476±0.020 & 0.428±0.020 \\
        \midrule
        \multirow{4}{*}{MLP Labels} 
            & Gpt4-0.5 & 0.831±0.010 & N/A & \textbf{0.609±0.009} \\
            & Gpt4-0.7 & 0.838±0.009 & N/A & 0.603±0.010 \\
            & Gpt4o-0.5 & 0.800±0.017 & N/A & 0.489±0.012 \\
            & Gpt4o-0.7 & 0.829±0.021 & N/A & 0.474±0.011 \\
       \midrule
        \multirow{4}{*}{Golden Labels} 
            & Gpt4-0.5 & 0.768±0.019 & 0.384±0.025 & 0.311±0.018 \\
            & Gpt4-0.7 & 0.770±0.013 & 0.391±0.021 & 0.332±0.020 \\
            & Gpt4o-0.5 & 0.778±0.020 & 0.384±0.016 & 0.287±0.085 \\
            & Gpt4o-0.7 & 0.813±0.029 & 0.395±0.028 & 0.338±0.026 \\
            \midrule
            
        \multicolumn{5}{l}{\textbf{MentalBERT}} \\
        \midrule
        \multirow{4}{*}{RLP Labels} 
            & Gpt4-0.5 & 0.821±0.005 & 0.533±0.009 & 0.548±0.017 \\
            & Gpt4-0.7 & \textbf{0.840±0.011} & \textbf{0.568±0.014} & 0.506±0.018 \\
            & Gpt4o-0.5 & 0.813±0.014 & 0.441±0.010 & 0.354±0.028 \\
            & Gpt4o-0.7 & 0.798±0.037 & 0.454±0.019 & 0.385±0.019 \\
        \midrule
        \multirow{4}{*}{MLP Labels} 
            & Gpt4-0.5 & 0.809±0.009 & N/A & \textbf{0.587±0.010} \\
            & Gpt4-0.7 & 0.828±0.015 & N/A & 0.570±0.007 \\
            & Gpt4o-0.5 & 0.791±0.009 & N/A & 0.472±0.003 \\
            & Gpt4o-0.7 & 0.811±0.023 & N/A & 0.420±0.017 \\
        \midrule
        \multirow{4}{*}{Golden Labels} 
            & Gpt4-0.5 & 0.772±0.014 & 0.367±0.011 & 0.320±0.018 \\
            & Gpt4-0.7 & 0.746±0.021 & 0.369±0.015 & 0.302±0.021 \\
            & Gpt4o-0.5 & 0.775±0.012 & 0.359±0.018 & 0.274±0.076 \\
            & Gpt4o-0.7 & 0.776±0.019 & 0.365±0.010 & 0.242±0.049\\
        
        \bottomrule
    \end{tabular}
    \caption{Mean Weighted F1 scores  with std on the test sets.}
    \label{tab:f1_weighted_scores_with_std}
\end{table*}

\section{CD Modeling Results}
\label{appendix:results}
Table \ref{tab:f1_weighted_scores_with_std} provides a comprehensive breakdown of the F1 (weighted) scores across all model and dataset configurations, classification tasks, and label sets considered in this research. This detailed table complements the main results presented in Table \ref{tab:f1_weighted_scores} by offering a granular view of performance differences between the MentalBERT and MentalRoBERTa models.

\section{Random Baseline ($F1_{random}$)}
\label{appendix:random f1}
Theoretical results in Table \ref{tab:random_f1_weighted_scores} present the random weighted F1 scores calculated based on the proposed formula in section \ref{sec:randomf1} across dataset configurations and classification tasks for the test set(s). We further verify these values through empirical testing.

\begin{table*}[h!]
\centering
\small
\begin{tabular}{llcccccc}
\toprule
\textbf{Label} & \textbf{Configuration} & \multicolumn{3}{c}{\textbf{Theoretical Results}} & \multicolumn{3}{c}{\textbf{Empirical Results (Mean ± Std)}} \\
\cmidrule(lr){3-5}
\cmidrule(lr){6-8}
 & & \textbf{Binary} & \textbf{Multiclass} & \textbf{Multilabel} & \textbf{Binary} & \textbf{Multiclass} & \textbf{Multilabel} \\
\midrule
\multirow{4}{*}{RLP} 
&Gpt4-0.5 & 0.589 & 0.196 & 0.348 & 0.588 ± 0.022 & 0.196 ± 0.019 & 0.347 ± 0.022 \\
&Gpt4-0.7 & 0.590 & 0.204 & 0.325 & 0.590 ± 0.023 & 0.204 ± 0.020 & 0.325 ± 0.021 \\
&Gpt4o-0.5 & 0.593 & 0.178 & 0.217 & 0.593 ± 0.022 & 0.177 ± 0.018 & 0.217 ± 0.020 \\
&Gpt4o-0.7 & 0.568 & 0.184 & 0.207 & 0.568 ± 0.022 & 0.185 ± 0.020 & 0.208 ± 0.020 \\
\midrule

\multirow{4}{*}{MLP} 
&Gpt4-0.5 & 0.569 & N/A & 0.455 & 0.570 ± 0.023 & N/A & 0.455 ± 0.026 \\
&Gpt4-0.7 & 0.576 & N/A & 0.413 & 0.577 ± 0.023 & N/A & 0.413 ± 0.024 \\
&Gpt4o-0.5 & 0.585 & N/A & 0.253 & 0.585 ± 0.021 & N/A & 0.254 ± 0.020 \\
&Gpt4o-0.7 & 0.568 & N/A & 0.238 & 0.568 ± 0.023 & N/A & 0.238 ± 0.022 \\
\midrule

\multirow{4}{*}{Golden} 
&Gpt4-0.5 & 0.538 & 0.176 & 0.201 & 0.538 ± 0.024 & 0.176 ± 0.016 & 0.200 ± 0.018 \\
&Gpt4-0.7 & 0.539 & 0.174 & 0.202 & 0.537 ± 0.024 & 0.175 ± 0.017 & 0.202 ± 0.018 \\
&Gpt4o-0.5 & 0.547 & 0.167 & 0.199 & 0.547 ± 0.024 & 0.166 ± 0.018 & 0.197 ± 0.019 \\
&Gpt4o-0.7 & 0.548 & 0.166 & 0.203 & 0.548 ± 0.025 & 0.166 ± 0.017 & 0.201 ± 0.019 \\
\bottomrule
\end{tabular}
\caption{Comparison of Theoretical (from \ref{sec:randomf1}) and Empirical Random F1 Weighted Scores Across Configurations and Classification Tasks for Test Set(s)}
\label{tab:random_f1_weighted_scores}
\end{table*}

\subsection{Empirical Verification}
To verify the accuracy and robustness of our proposed random F1 score calculation methodology, we performed empirical tests by simulation random label assignment using \texttt{np.random.choice} method in Python. The verification process involved generating random predictions based on the observed class distributions for each dataset and label type, which were treated as model predictions for calculating the corresponding weighted F1 scores. This process is repeated 1000 times for each configuration, and the averaged weighted F1 values are reported in Table \ref{tab:random_f1_weighted_scores} along with the corresponding standard deviation.

Table \ref{tab:random_f1_weighted_scores} shows empirical results closely matched with the theoretical scores, with deviations well within the expected standard error, thus validating our approach. The close alignment between the empirical and theoretical random F1 scores serves as strong evidence that our random F1 score calculation methodology is both accurate and reliable. The verification process also reinforces the validity of using these baseline scores in our kappa calculation ($\kappa_{F1}$), enabling a fair comparison of model performance across diverse datasets and label types.

\section{Human Verification}
\label{appendix:sampling}

\paragraph{Sampling Strategy:} We employed a structured sampling strategy for manual verification. This approach aimed to cover the diversity of labels within our dataset while maintaining a fair representation of both frequent and rare labels. We selected annotations generated using the GPT4-0.5 dataset for this process. To construct a well-balanced and meaningful sample for manual verification, we followed a multi-step process:

\begin{enumerate}
\item \textbf{Subsetting Based on RLP and MLP Labels:} We created a subset where all labels in the \textsc{RLP Labels} set were present in the \textsc{MLP Labels} set, retaining 75.42\% of the GPT4-0.5 dataset.
\item \textbf{Strict Mismatch Subset Creation:} From the above subset, we identified instances where \textsc{RLP Labels} and \textsc{MLP Labels} had no overlap with the \textit{golden labels}, achieving a strict mismatch subset of 35.61\% of the original GPT4-0.5 dataset.
\item \textbf{Filtering by Maximum Repetition:} To enhance label reliability, we filtered for instances with a maximum label repetition of 5, yielding 31.54\% of the original GPT4-0.5 dataset.
\item \textbf{Final Sampling Approach:} All instances containing less frequent labels were included entirely, while more frequent labels were capped at 10 instances each. This approach resulted in a final sample size of 101 instances i.e, 3.99\% of the original GPT4-0.5 dataset, ensuring a manageable yet meaningful sample set.
\end{enumerate}

\paragraph{Verification Process:} We utilized \textit{Label Studio} for conducting the manual verification of our annotated labels. In this process, 3 domain experts were presented with a randomized sample where each instance included:
\begin{itemize}
\item \textbf{Original Text:} Providing context (\textsc{user input}) for accurate assessment.
\item \textbf{Label 1} and \textbf{Label 2:} These Labels were shuffled randomly to maintain blinding, ensuring experts could not trace which Label was generated by LLM and which originated from the golden standard.
\end{itemize}

Experts were asked to select which label they agree with using the following options:
a) \textbf{Label 1}, b) \textbf{Label 2}, c) \textbf{Both Labels}, d) \textbf{None of the Labels}, e) \textbf{Partial Label 1} (if applicable), and f) \textbf{Partial Label 2} (if applicable). This blinded evaluation method promotes unbiased feedback, allowing us to assess the quality of our model's annotations against expert judgment.

\end{document}